\documentclass[conference]{IEEEtran}

\IEEEoverridecommandlockouts
\usepackage{cite}
\usepackage{amsmath,amssymb,amsfonts}
\usepackage{algorithmic}
\usepackage{graphicx}
\usepackage{textcomp}
\usepackage{xcolor}
\usepackage{url}
\usepackage{etoolbox}
\makeatletter
\patchcmd{\thebibliography}{\footnotesize}{\scriptsize}{}{}
\patchcmd{\@IEEEthebibliography}{\footnotesize}{\scriptsize}{}{}
\makeatother

\def\BibTeX{{\rm B\kern-.05em{\sc i\kern-.025em b}\kern-.08em
    T\kern-.1667em\lower.7ex\hbox{E}\kern-.125emX}}

\makeatletter
\def\@maketitle{%
  \newpage
  \null
  \vskip 10mm%
  \begin{center}%
    \let \footnote \thanks
    {\LARGE \bfseries \@title \par}%
    \vskip 1.5em%
    {\large
      \lineskip .5em%
      \begin{tabular}[t]{c}%
        \@author
      \end{tabular}\par}%
  \end{center}%
  \par
  \vskip 1.5em}
\makeatother
\begin{document}

\title{3D Printing of Passively Actuated Self-Folding Robots with Integrated Functional Modules}

\author{
\IEEEauthorblockN{
Gaolin Ge\IEEEauthorrefmark{1}$^{1}$,
Qifeng Yang\IEEEauthorrefmark{1}$^{2}$,
Haoran Lu$^{1}$,
Tingyu Cheng$^{4}$,
Martin Nisser\IEEEauthorrefmark{2}$^{3}$ and Yiyue Luo\IEEEauthorrefmark{2}$^{2}$
}

\thanks{\scriptsize\IEEEauthorrefmark{1} These authors contributed equally.}
\thanks{\scriptsize\IEEEauthorrefmark{2} Co-corresponding authors. Email: nisser@uw.edu, yiyueluo@uw.edu}

\thanks{\scriptsize
$^{1}$Mechanical Engineering, University of Washington, Seattle, WA, USA. 
$^{2}$Electrical \& Computer Engineering, University of Washington, Seattle, WA, USA. 
$^{3}$William E. Boeing Department of Aeronautics \& Astronautics, University of Washington, Seattle, WA, USA. 
$^{4}$Computer Science and Engineering, University of Notre Dame, Notre Dame, IN, USA.}
}

\maketitle

\begin{abstract}
We introduce an elastic-driven self-folding approach that fabricates robots directly from flat 3D-printed conductive PLA nets. Elastic bands routed through printed hooks store energy that folds the sheet into programmed 3D geometries, while the flat state allows accurate placement of electronics and magnets before deployment. The same substrate doubles as electrodes for capacitive touch and supports a reusable platform I/O palette with Hall sensors and eccentric rotating mass (ERM) motors for docking detection and vibration actuation. We also derive a closed-form folding model that balances hinge stiffness with elastic band moment to predict equilibrium fold angles; experiments validate the model and yield a design map linking hinge thickness, band size, and hook spacing to target angles. Using this workflow we realize multiple polyhedral modules and demonstrate three applications: a cube that highlights the potential of self-folding for scalable modular robot collectives, a deployable gripper, and a tendon-driven finger. The method is low cost, stimulus-free, and integrates actuation and sensing.
\end{abstract}


\section{Introduction}
Robots are becoming increasingly affordable and accessible, which is driving new opportunities in education, research, and large-scale collective systems \cite{Atman2023,  Ryalat2025, Calderon2022}. In these contexts, it is especially important to design robots that can be produced rapidly and at low cost, so that large numbers of units can be fabricated without extensive manual effort. Self-folding robotics has emerged as a promising approach for this challenge, allowing planar structures to transform into functional three-dimensional geometries. 

Prior self-folding methods based on shape-memory laminates or specialized composites require heat, custom materials, or multilayer processes, which limits accessibility and repeatability \cite{Tolley2013, Felton2013soft, Felton2013, Felton2014, niu2023pullupstructs}. Directly printing hollow 3D bodies avoids special materials but makes the placement of electronics and sensors difficult once interiors are enclosed \cite{Miyashita2015}. Other approaches place actuation or sensing directly at the creases. For example, pneumatic pouch motors \cite{Sun2015ICRA} can be embedded along fold lines to enable reversible, pressure-driven folding, and inkjet-printed angle sensors \cite{Sun2015IROS} have been integrated for closed-loop fold regulation. However, these systems typically rely on laminate stacks, external pressure sources, and post-fold wiring, which increases tooling overhead and complicates scaling to electronics-ready batches \cite{Sun2015ICRA, Sun2015IROS}. 

We address this gap by introducing a design for self-folding robots that can be 3D-printed in a flat configuration, installed with electronic components, and folded into programmed target shapes using inexpensive elastic bands. The key idea is an elastic-driven folding mechanism that routes bands around hooks printed onto the robot body's pre-folded faces, producing folding without heat or specialized materials. The flat state also enables precise and rapid placement of electronics and magnets before deployment. Each robot is printed in conductive PLA as a planar net with compliant hinges, hooks, locks, and seats for components. Because conductive PLA is electrically active, the same printed surfaces can serve as capacitive sensing electrodes once folded, turning structural faces into touch-sensitive inputs. Combined with magnets and Hall sensors for docking detection, this allows modules to both act and sense without requiring additional wiring or separate sensor assemblies.

This paper makes four contributions:
\begin{enumerate}
    \item An accessible and low-cost self-folding method that uses elastic bands with printed hooks and thin hinges to fold flat nets into 3D shapes without heat or special materials.
    \item A model that links geometry and band parameters to the final fold angle, together with a characterization procedure that yields a practical design map for targeted angles.
    \item An assembly-while-flat workflow that prints structure, seating features, and conductive electrodes in one step, enabling reuse of common I/O components (ERM motors, magnets with Hall sensors, and conductive electrodes) across designs.
    \item Demonstrations across three robot platforms: a cubic swarm robot with vibration-based locomotion, touch-based command, and magnetic docking; a tendon-driven finger; and a foldable gripper.   
\end{enumerate}


\begin{figure*}[tb]
  \centering
  \includegraphics[width=\textwidth]{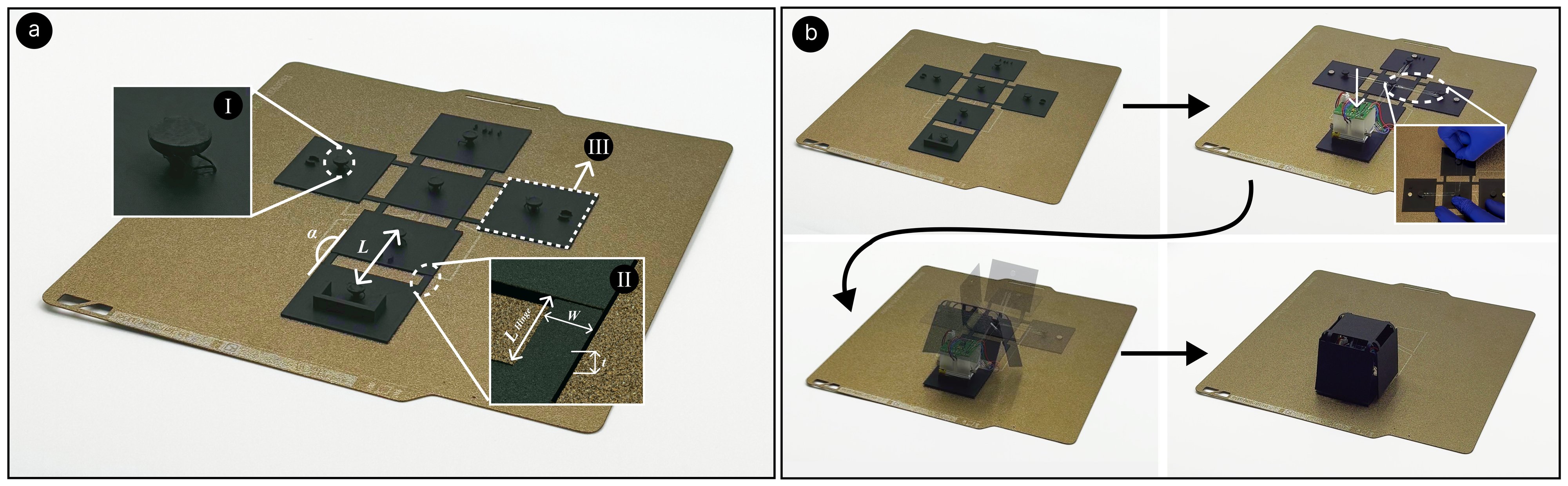}
  \caption{Mechanical overview of the folding process. (a) Printed conductive PLA net for a cube module with integrated components: I. hook, II. hinge, and III. substrate. Geometric parameters are labeled in the inset. (b) Folding demonstration: after inserting elastic bands and placing electronics, the sheet is detached from the print bed and self-folds into a cube.}
  \label{overall workflow}
\end{figure*}

\section{Related Work}
Self-folding robotics has emerged as a promising approach for achieving rapid and easy fabrication of robotic systems, enabling the transition from simple planar structures into complex 3D geometries. Seminal work on self-folding robots utilized shape-memory laminates actuated by stimuli such as heat, electricity, and light to form predefined geometric shapes, including cubes and other polyhedral forms \cite{Tolley2013, Felton2013soft}. Progress in this area led to shape memory polymers (SMPs) combined with embedded resistive circuits, enabling precise self-folding into robotic structures \cite{Felton2013, Felton2014}. Complementing embedded resistive heating, Liu et al. introduced a wireless magnetic-induction route in which alternating fields locally heat conductive layers to trigger SMP creases, enabling remote folding through non-magnetic obstacles while avoiding electronics \cite{Liu2022IROS}. To further improve folding accuracy, feedback-controlled self-folding methods employing integrated sensors have been introduced, allowing robots to monitor and precisely control fold angles. This enhancement significantly advanced the precision and repeatability of self-assembly in robotic collectives \cite{Nisser2016}.

Swarm and modular robots often require large numbers of inexpensive units to demonstrate collective behaviors. Prior work shows that collectives such as Kilobot scale to thousands of units by minimizing per-robot hardware \cite{Rubenstein2012}, while modular platforms like ModQuad highlight the role of structured docking in building large assemblies \cite{Brandt2018ModQuad, Han2023IROS}. Cheap and rapid self-folding provides a complementary path toward this goal, enabling the fabrication of modules or swarm-capable robots with low part count and integrated features.

Inspired by these foundational studies, our work introduces an elastic-driven self-folding mechanism that employs elastic bands integrated with 3D-printed conductive PLA structures. In contrast to prior approaches relying on specialized materials or external stimuli, our method offers a resettable and stimulus-free folding process with low cost, widely available elastic bands, an assembly-while-flat approach for integrating electronics, and a design map linking geometry to fold parameters. These features enable scalable fabrication and practical deployment of foldable robotic modules with built-in sensing and locomotion, making the approach particularly well-suited for swarm robotics and deployable manipulators.

\section{Method}

\subsection{System Design}

The folding mechanism is implemented on a single 3D-printed conductive PLA sheet patterned as a polyhedral net. Prints were fabricated on a Bambu Lab P1S using Protopasta Electrically Conductive Composite Black PLA filament. This planar substrate integrates structure, actuation, and sensing in one layer, enabling rapid prototyping and direct integration of mechanical and electronic features (Fig.~\ref{overall workflow}a).

\subsubsection{Hooks}
Hooks are small printed posts designed to guide the elastic bands along predetermined paths and to hold them in place during folding. Their placement defines the initial pre-strain in the bands, which directly sets the folding force applied to each hinge.

\subsubsection{Hinges}
Local thin regions in the substrate act as compliant hinges. Their thickness and width determine bending stiffness and durability. By co-printing these regions with the substrate, the design eliminates separate hinge assembly and provides consistent folding performance.

\subsubsection{Substrate}
The substrate is the flat-printed sheet that forms the base of the foldable robot. Conductive PLA gives both mechanical support and electrode functionality, so the same sheet serves as chassis and sensor. The flat geometry simplifies fabrication and allows seats, hooks, and hinges for electronic components to be incorporated directly.

Together, these three elements create a simple but effective folding system. Hooks guide the elastic bands and keep the folding force consistent, hinges allow smooth bending without the need for extra parts, and the flat sheet combines both structural and sensing functions. This approach reduces the number of parts, makes assembly easier, shortens fabrication time, and keeps all features printed in a single layer.

\subsection{Model for Self-folding}


To model the folding mechanism, the system is treated as the interaction between two antagonistic springs: a torsional spring (the hinge) and a linear spring (the elastic band). The equilibrium fold angle emerges from the balance of these opposing moments. For small deformations, elastic-band elastomers can be treated as linear Hookean springs ~\cite{Wu2025}; consequently, an elastic band spanning a hinge is modeled as two symmetric linear springs (each of stiffness $k_b$) in parallel \cite{Deng2022}.

The parameters related to folding, shown by Fig.~\ref{overall workflow}a, include hinge thickness $t$, hinge width $W$, hinge length $L_{\text{hinge}}$, folding hinge spring constant $k_h$, internal diameter of the relaxed elastic band loop $d$, elastic band spring constant $k_b$, hook spacing $L$, and equilibrium fold angle $\alpha$.   

\subsubsection{Geometric relation}  
For symmetric hook placement, the hook-to-hook distance is expressed as  
\begin{equation}
D(\alpha) = L \sin\left(\tfrac{\alpha}{2}\right).
\end{equation}
The elastic band extension is therefore  
\begin{equation}
\Delta \ell(\alpha) = D(\alpha) - L_0;
\end{equation}

\subsubsection{Elastic band spring model}
One elastic band, when attached to hooks, is modeled as two identical linear springs trying to bend the hinge. The effective relaxed length between two hooks can be approximated as a fraction of the elastic band’s circumference. Because the band wraps partially around the hook posts, the relaxed path is shorter than half of its free circumference. We therefore write

\begin{equation}
L_0 = \gamma \, \pi \frac{d}{2},
\end{equation}

where $d$ is the internal diameter of the elastic band loop and $L_0$ is the effective relaxed length between hooks. The coefficient $\gamma \approx 0.9$ accounts for the reduction due to finite hook diameter and contact curvature, and was obtained empirically from repeated measurements.

The elastic force generated by the stretched band is then
\begin{equation}
F_b(\alpha) = 2 k_b \, \Delta \ell(\alpha),
\end{equation}
and the corresponding moment about the hinge is
\begin{equation}
M_b(\alpha) = F_b(\alpha)\, \frac{L}{2}\cos\left(\tfrac{\alpha}{2}\right).
\end{equation}

\subsubsection{Hinge torsional spring model}  
The hinge is modeled as a torsional spring. Following Euler–Bernoulli beam theory, the hinge behaves like a cantilever beam. Although this theory is derived for small deflections, Wagner’s experiments \cite{Wagner2020} demonstrate that 3D-printed hinges behave similarly to this model even under large bending. The torsional stiffness is 
\begin{equation}
k_h = \frac{E W t^3}{12 L_{\text{hinge}}},
\end{equation}
where $E$ is the Young’s modulus of PLA. 
\begin{equation}M_h(\alpha) = k_h (\pi - \alpha).
\end{equation}

\subsubsection{Equilibrium condition}  
The equilibrium fold angle $\alpha$ is obtained when the hinge and elastic band moments are balanced:  
\begin{equation}
M_h(\alpha) = M_b(\alpha).
\end{equation}
This relation defines the mapping from geometric and material parameters to the equilibrium fold angle. Such a mapping allows designers to quickly predict folding outcomes without trial-and-error prototyping, making it easier to select hinge thickness, band size, and hook spacing to achieve a desired target angle.



\subsection{Flat Assembly and Platform I/O}
The flat-printed sheet works as both the robot frame and a template for placing parts. As shown in Fig.~\ref{overall workflow}b, all components are mounted into printed seats and slots while the sheet is still flat, which keeps everything aligned. When the sheet is scraped off the build plate, the pre-routed elastic bands drive automatic folding and enclose the parts in their intended positions. This workflow eliminates the need to insert components into closed cavities and reduces assembly time.

Viewed across modules, this workflow defines a shared set of input and output parts installed in the flat stage: ERM motors for vibration-based locomotion, press-fit magnets for passive latching and alignment, linear Hall-effect sensors for detecting magnetic contacts, and printed conductive faces that act as capacitive electrodes. We characterize these interfaces once and reuse the same mounts, wiring, and thresholds across applications. We reuse the same controller PCB across modules without redesign. For sensing, we reuse the same fixed thresholds, and calibration is limited to a short baseline capture at start.

\subsection{Sensorizing the Structure with Conductive PLA}

Beyond the integration of additional modules, functionality can also be embedded directly through the choice of printing materials. In our case, conductive PLA provides both structural support and electrical activity, allowing the same printed sheet to act as electrodes for capacitive touch sensing \cite{Schouten2021}. The flat-printed sheet therefore doubles as an electrode array: when connected to an MPR121 capacitive sensor \cite{MPR1212010}, a user’s touch on a face produces a measurable change in capacitance. The main controller detects this change as a touch event, enabling direct human–robot interaction through the robot’s outer surfaces.

\section{Results}
In this section we validate the proposed self-folding model and demonstrate how the same process of elastic-driven folding, flat assembly, and sensing through the printed structure can be applied across different modules. We first characterize folding mechanics and the design map, then present three application examples: a swarm robot, a tendon-driven finger, and a deployable gripper.

\subsection{Self-folding Characterization}
We conducted folding tests on PLA structures to validate the self-folding mechanism and model. Flat PLA coupons and full polyhedral net prototypes were printed with systematic variations in one design parameter at a time. The design space includes (i) hinge thicknesses, varied to adjust bending stiffness, and (ii) elastic band pre-tensions, realized using bands of different dimensions and stiffness. In particular, three elastic band types were tested: yellow bands with diameter $d=12\text{mm}$ and thickness $t=1.4\text{mm}$, transparent bands with $d=15\text{mm}$ and $t=2.0\text{mm}$, and black bands with $d=25\text{mm}$ and $t=1.4\text{mm}$. After inserting the elastic bands, each sample was released from the flat state and allowed to self-fold under the bands’ elastic tension. We measured the resulting equilibrium fold angle for each configuration and repeated each experiment three times to ensure consistency. This procedure produced a dataset of achieved fold angles across the parameter space for comparison with the model predictions.

\begin{figure}[ht]
  \centering
  \includegraphics[width=\columnwidth]{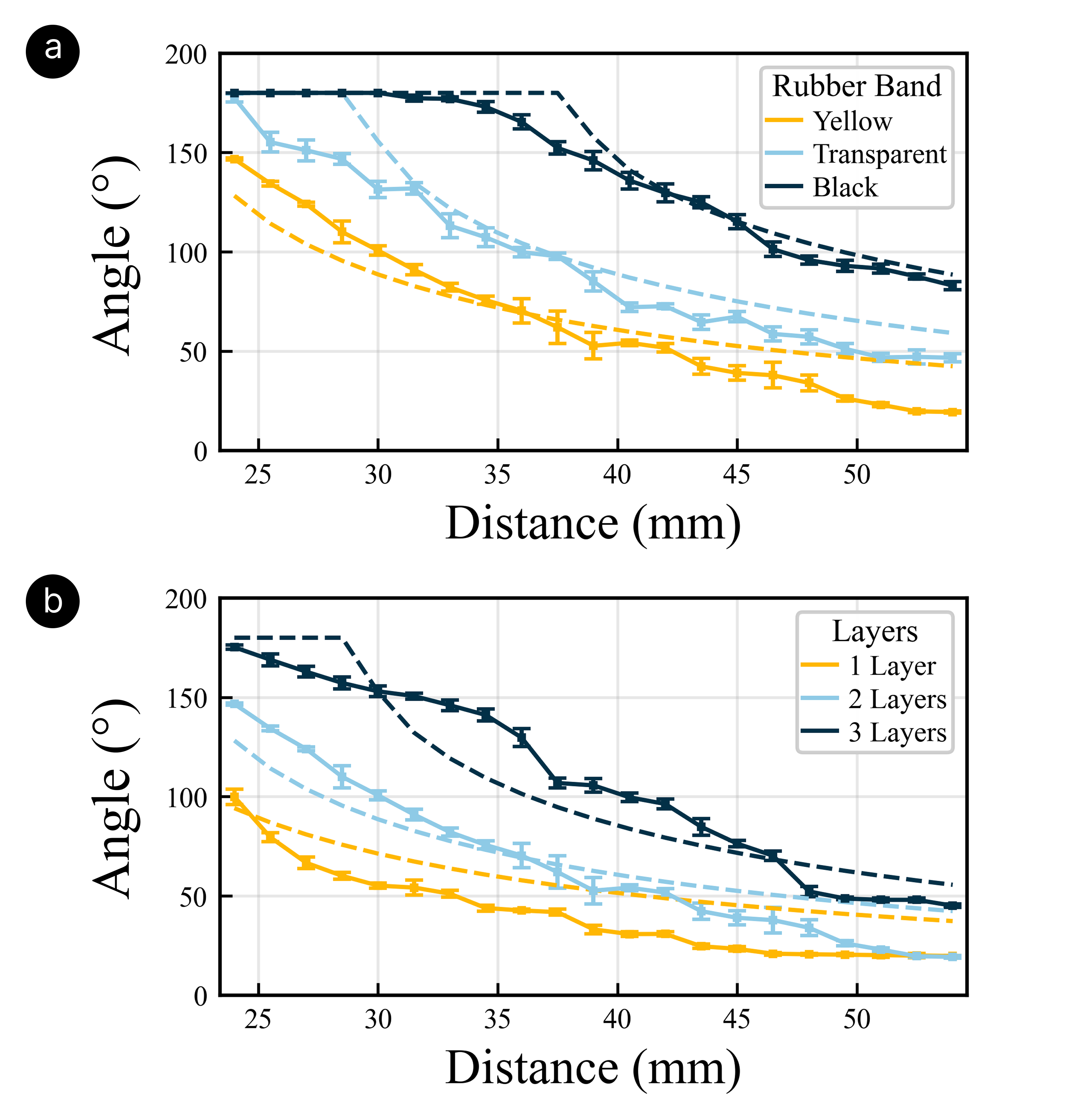}
  \caption{Folding characterization results. (a) Equilibrium fold angles versus hook spacing for three elastic band types (yellow: $d=12\text{mm}$, $t=1.4\text{mm}$; transparent: $d=15\text{mm}$, $t=2.0\text{mm}$; black: $d=25\text{mm}$, $t=1.4\text{mm}$). (b) Equilibrium fold angles versus hook spacing for hinges with one, two, and three printed layers. Solid lines are experimental means with error bars; dashed lines are model predictions.}
  \label{fig:LayerandBand}
\end{figure}

As shown in Fig.~\ref{fig:LayerandBand}, the measured fold angles generally agree with the model across elastic band types and hinge thicknesses, with a small deviation for the one-layer hinge condition. This discrepancy is likely due to first-layer printing effects: the printer applies different settings on the initial layer, which locally increase the effective hinge thickness and stiffness and shift the equilibrium angle \cite{Wagner2020, Grgic2023, Ramful2024, Mutlu2016}. With the model validated, we generated a design map linking target fold angle $\alpha$ to hinge thickness, band size, and hook spacing. In practice, a designer can choose these parameters from the map to realize a desired $\alpha$. Using this map, we deliberately folded flat nets into predefined 3D polyhedra.

\begin{figure}[ht]
  \centering
  \includegraphics[width=\columnwidth]{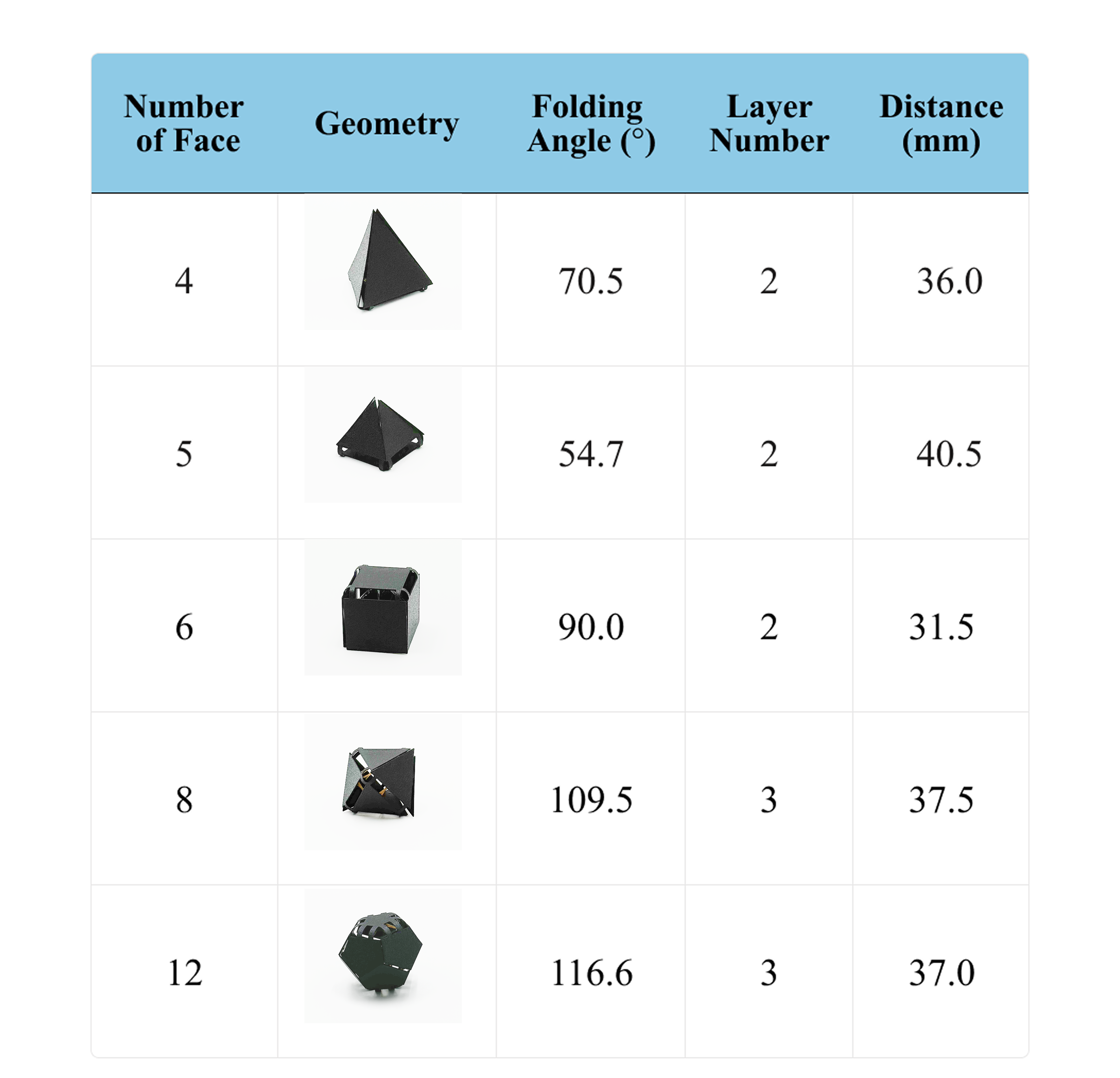} 
  \caption{Parameters of foldable polyhedral modules. Shown are representative models with different face counts, along with their corresponding target dihedral folding angles (°), hinge layer numbers, and elastic-band anchor spacings (mm).}
  \label{fig: PolyTable}
\end{figure}

As shown in Fig.~\ref{fig: PolyTable}, by tuning parameters via the map we realized a tetrahedron, square pyramid, cube, octahedron, and dodecahedron \cite{Sayapin2019}.

\subsection{Swarm Robot}
This example applies our approach to a cube module for swarm studies, highlighting that elastic self-folding with flat assembly and printed electrodes enables low-cost, repeatable integration of locomotion, docking, and touch.

\subsubsection{Self-folding}
\begin{figure}[ht]
  \centering
  \includegraphics[width=\columnwidth]{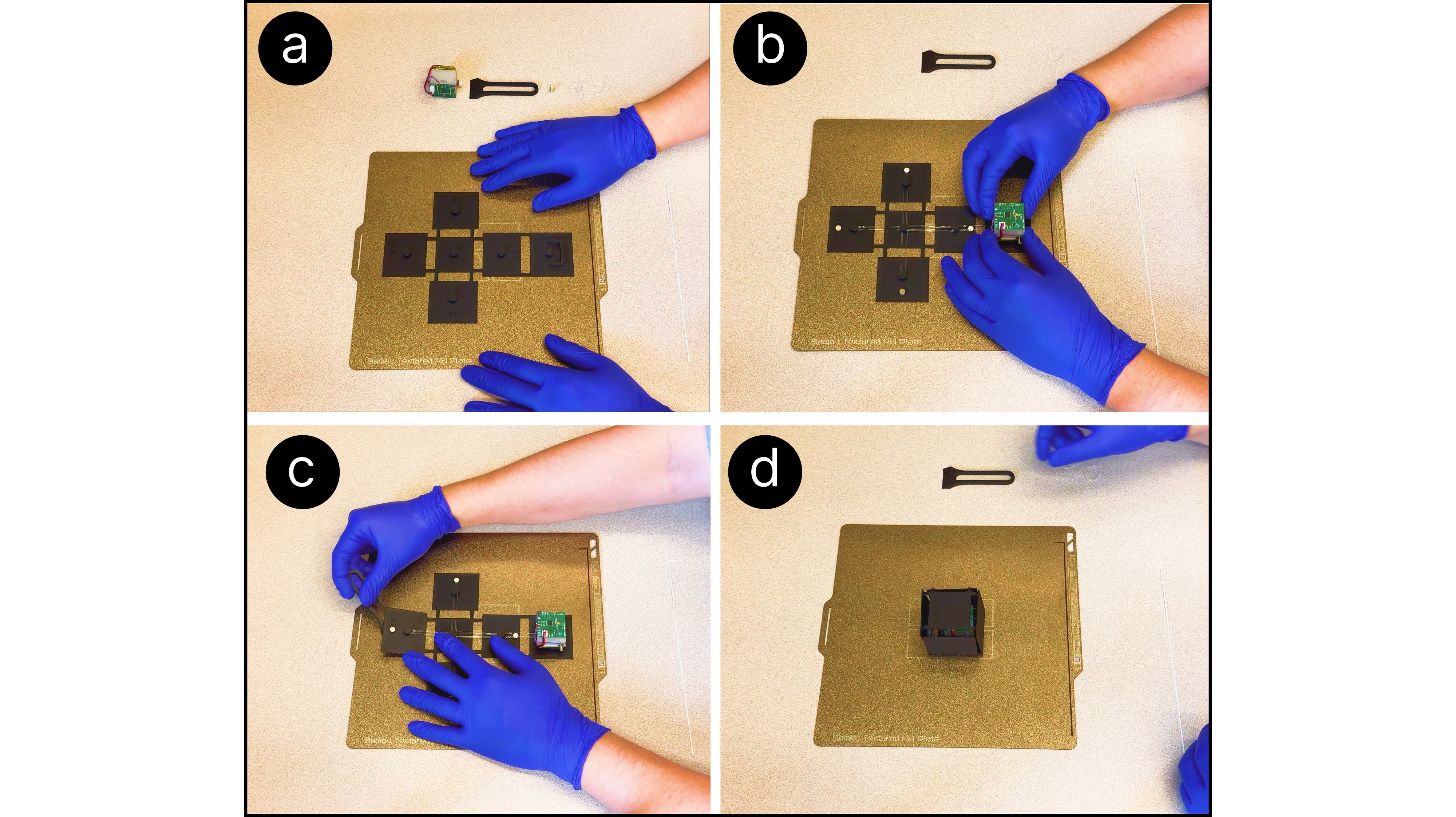}
  \caption{Assembly process of the foldable swarm robot. (a) Layout of all components. (b) Placement of elastic bands, magnets, sensing unit, and PCB with ERM motors onto the flat substrate. (c) Detachment of the conductive PLA sheet from the print bed. (d) Final self-folded cube module.}
  \label{fig:SWrobot}
\end{figure}

The cubic swarm robot demonstrates how elastic-driven self-folding with conductive PLA preserves the alignment set during flat assembly. As shown in Fig.~\ref{fig:SWrobot}, the module is printed as a flat pattern with hooks, hinges, sensor seats, magnet slots, and electrode pads. Once scraped off the build plate, pre-routed elastic bands fold the sheet into a cube, the most common geometry in modular swarm systems \cite{Yim2007}. Folding encloses the components and locks them in place.

\subsubsection{Flat assembly}
ERM vibration motors form part of the platform I/O integrated during the assembly-while-flat workflow. They are embedded in adapters that connect directly to the PCB and substrate in the flat state. To characterize this channel for reuse across modules, we measured motion response as a function of PWM duty. Overhead videos were recorded with a fixed camera, and an OpenCV tracker extracted the robot's position frame by frame \cite{OpenCVLibrary}. The sequence of extracted positions was then used to generate the robot’s trajectory under different PWM duty cycles.

\begin{figure}[ht]
  \centering
  \includegraphics[width=\columnwidth]{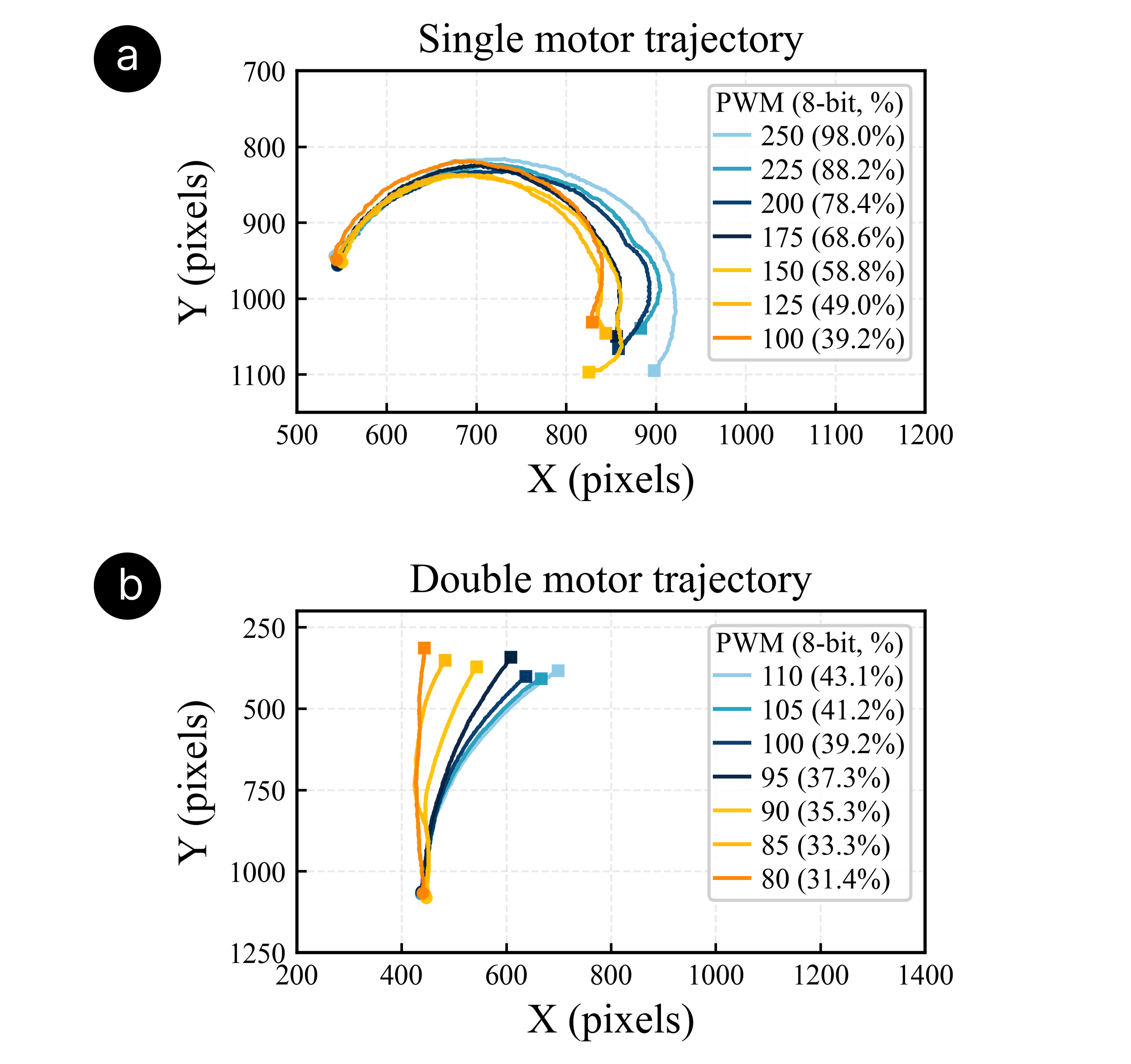}
  \caption{Module trajectories under ERM actuation. (a) Single-motor: actuation alone produces circular arcs; higher duty (39.2 to 98.0\%) reduces curvature and increases forward displacement. (b) Dual-motor: equal drive yields curved motion due to mass offset; reducing one motor from 43.1\% to 31.4\% straightens the path.}
  \label{fig:SRCharTra}
\end{figure}

\textit{Single-motor test.} With one motor actuated and the other off, the module follows a circular trajectory. As the duty cycle increases from 39.2\% to 98.0\%, the path curvature decreases (Fig.~\ref{fig:SRCharTra}a). This trend matches ERM dynamics: higher motor speed increases inertial excitation
\[
F_c = m r \omega^{2},
\]
which enlarges the slip-stick impulse per cycle and yields greater net forward displacement on flat surfaces \cite{Rubenstein2012, Vartholomeos2006, Majewski2017, SimoBot2022}.

\textit{Dual-motor test.} When both motors run at the same duty, the module exhibits a slight curvature because the center of mass is not perfectly centered in the folded cube. To compensate, one motor was fixed at 43.1\% duty while the other was reduced from 43.1\% to 31.4\%. The trajectory transitioned from curved to nearly straight as the second motor’s duty decreased, demonstrating that fine-tuning of relative motor power can overcome mass imbalance and produce controllable straight-line motion (Fig.~\ref{fig:SRCharTra}b).

\subsubsection{Sensing via structure}
\begin{figure}[ht]
  \centering
  \includegraphics[width=\columnwidth]{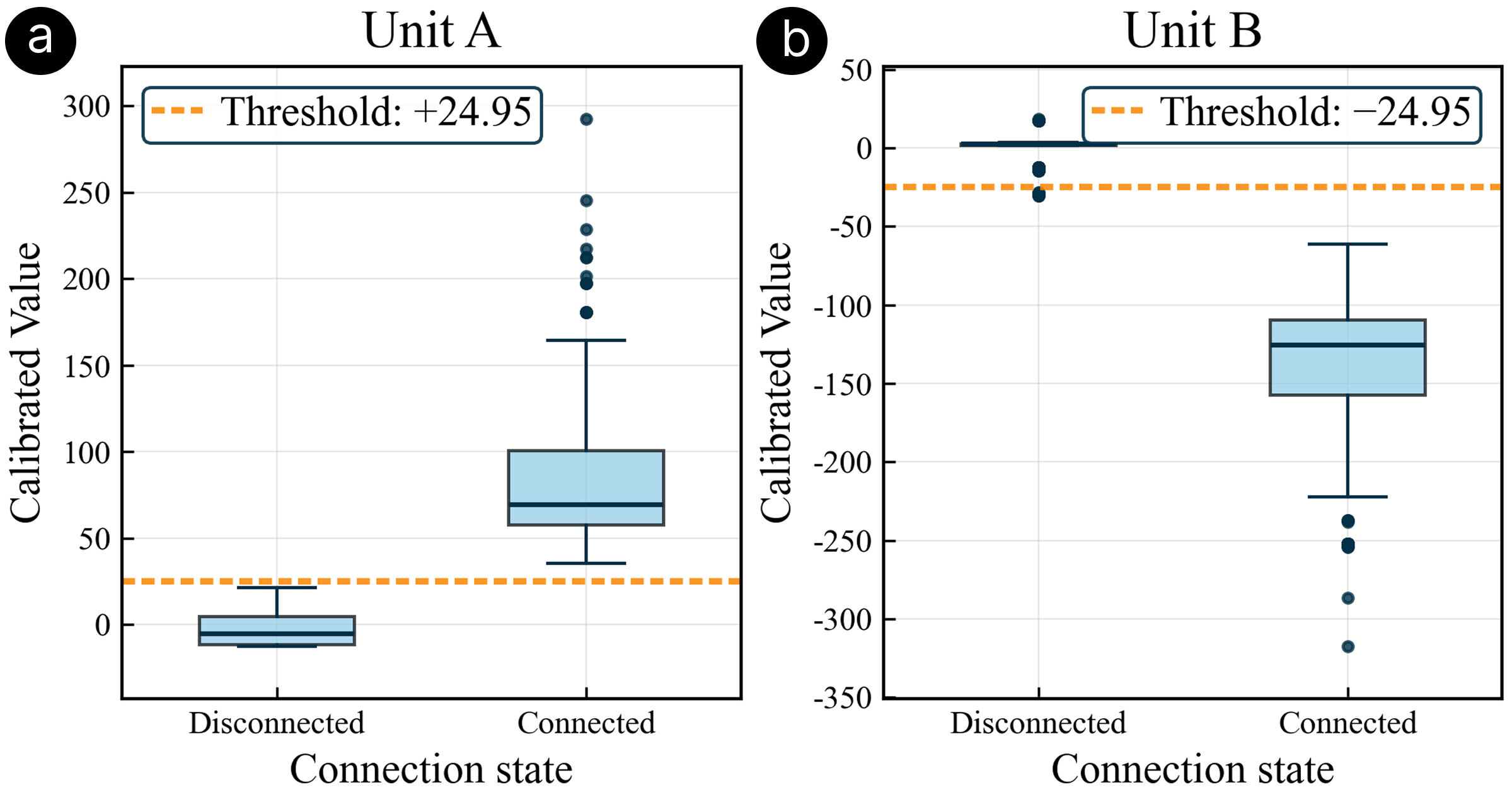}
  \caption{Magnet sensing results. Hall-effect readings show distinct connected and disconnected states, with a fixed threshold ($24.95$) enabling reliable classification despite opposite sensor orientations. Magnet sensing results for Units A and B in connected and disconnected states. (a) Hall sensor readings measured on Unit A. (b) Hall sensor readings measured on Unit B.}
  \label{fig:SRCharMag}
\end{figure}

Magnets and Hall-effect sensors form a paired I/O channel: magnets provide passive latching and alignment, while Hall sensors report the magnetic state. Both are installed during flat assembly, with sensors press-fit into co-printed seats and magnets inserted into aligned slots, so that their relative orientation is fixed before folding. After folding, linear Hall sensors (SS49E) measure local magnetic flux density across cube faces to monitor docking \cite{Honeywell2015}. To establish a threshold, we executed repeated connect–disconnect cycles, subtracting a baseline from the first 20 disconnected samples. The SS49E outputs raw ADC counts (unitless), with $12$ bits$0$ to $4095$ counts, digitized by the ESP32-C6 ADC. As shown in Fig.~\ref{fig:SRCharMag}, unit A measured $-3.676 \pm 7.602$ (disconnected) and $79.504 \pm 34.666$ (connected), while unit B measured $-0.152 \pm 6.744$ and $-134.125 \pm 40.758$. The opposite signs come from sensor orientation. Using the magnitude of the change, we set a fixed threshold of $24.954$, above which docking is detected. To quantify noise robustness beyond thresholding, we computed the signal-to-noise ratio (SNR) between the connected and disconnected Hall states using the within-state fluctuations as noise. Across experiment, we achieved an average SNR of $18.61 \pm 5.74\text{dB}$, indicating a clear margin for reliable docking classification.

\begin{figure}[ht]
  \centering
  \includegraphics[width=\columnwidth]{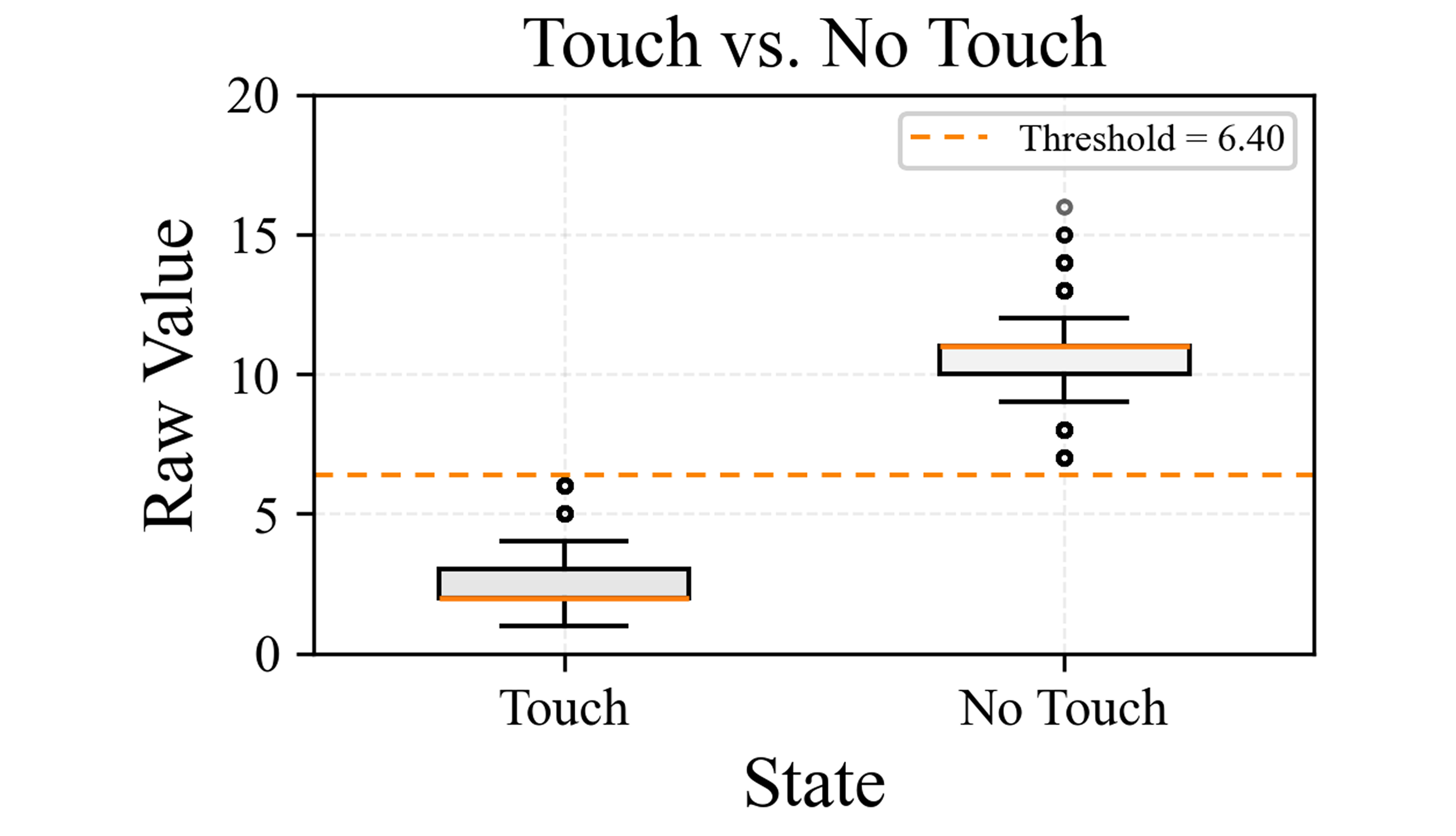}
  \caption{MPR121 touch sensing characterization, showing separation between touch and no touch at a threshold of 6.40.}
  \label{fig:MPRchar}
\end{figure}

Because the sheet is printed in conductive PLA, electrode pads from the flat design become capacitive faces after folding. We tested the touch interface to set a threshold and check responsiveness. In calibration, MPR121 readings averaged $10.57 \pm 0.82$ (no touch) and $2.23 \pm 0.94$ (touch). A single threshold of 6.4 separated the two distributions (Fig.~\ref{fig:MPRchar}). 

\begin{figure}[ht]
  \centering
  \includegraphics[width=\columnwidth]{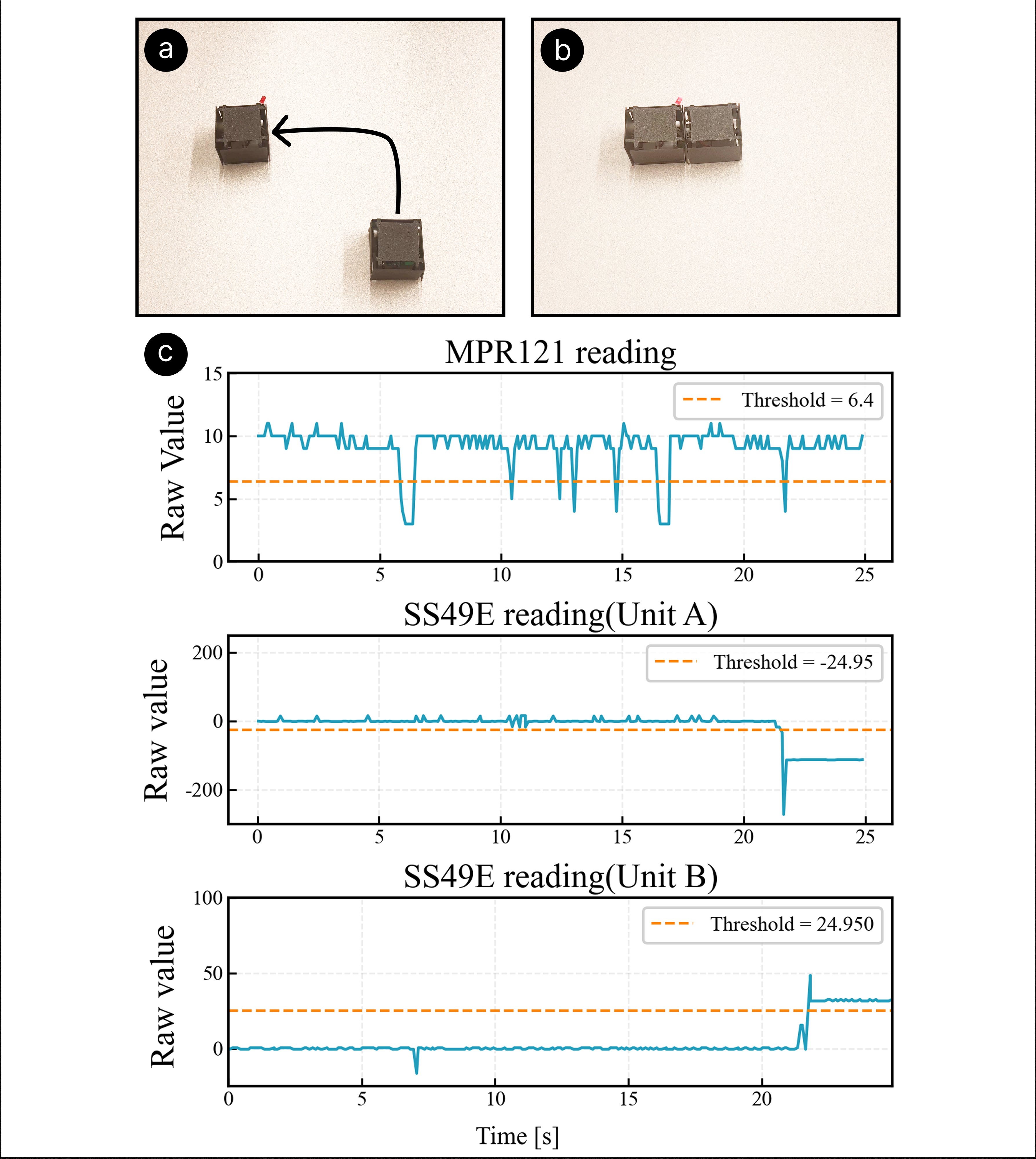}
  \caption{Swarm robot with integrated sensing and control. (a-b) Unit A approaches and docks with Unit B following a touch-triggered L-shaped trajectory. (c) Sensor readings from the MPR121 capacitive sensor and SS49E Hall sensors on Units A and B, with thresholds indicated, showing reliable detection of motion commands and docking events.}
  \label{fig:demoSW}
\end{figure}

Robustness was confirmed by tapping at 50–200 BPM for 3 min each; accuracy stayed above 98\% (99.3\% at 50 BPM, 98.0\% at 150–200 BPM). Using these channels, Unit A responded to touch commands for forward motion, turning, and stopping, then docked with Unit B; Hall sensing confirmed the connection and triggered an LED (Fig.~\ref{fig:demoSW}). We further report an SNR metric to quantify separation between touch and no-touch distributions. Using the within-state variation as noise, the MPR121 touch channel achieved a mean SNR of $20.06 \pm 0.78\text{dB}$ across trials, supporting robust touch event detection.

\subsection{Finger Module}
This example extends the approach to a tendon-driven finger, showing that the same self-folding and flat assembly pipeline supports servo actuation and printed touch input with minimal added parts.

\subsubsection{Self-folding}
This application presents a tendon-actuated finger built with a flat-to-folded fabrication method (Fig.~\ref{fig: FingerDemoData}). The design follows cable-routed fingers (e.g., \cite{Xstrings2025, Tanaka2023}) but replaces the rigid cable with a single elastic band. The finger is printed as a planar net and then folded into its three-dimensional form. Three flexure hinges connect the phalanges, and co-printed mechanical stops bound the joint ranges. Because one micro-servo drives all three joints, these stops define the closing sequence and final pose while protecting the hinges from over-bending \cite{krut2005, laliberte2002, Feshbach2023IROS}. During actuation, servo rotation increases tendon tension, generating folding moments at the joints until each contacts its stop. Stops were designed for 120°, 90°, and 100° at the base, middle, and distal joints; measurements confirmed 120°, 90°, and 98°, respectively (Fig.~\ref{fig: FingerDemoData}e). When the servo unwinds, relaxed tendon together with hinge elasticity biases the finger back toward the extended pose; in our prototype an antagonistic tendon was not required \cite{catalano2014}.

\subsubsection{Flat assembly}
The tendon anchors at the fingertip, passes through low-friction guides, and winds on a servo spool in the palm. Path length, guide spacing, and hinge stiffness are tuned so the finger closes in a predictable order to repeatable end poses set by the stops. Relative to a traditional cable-driven finger, the elastic-tendon layout offers practical benefits within this workflow: fewer parts and simpler routing, tolerance to minor misalignment due to compliance, quick tendon replacement, and reduced sensitivity to precise pre-tension \cite{ma2013, massa2002}. These choices align with the flat-to-folded approach, where joints, stops, and routing are integrated on the printed sheet.

\subsubsection{Sensing via structure}
\begin{figure}[ht]
  \centering
  \includegraphics[width=\columnwidth]{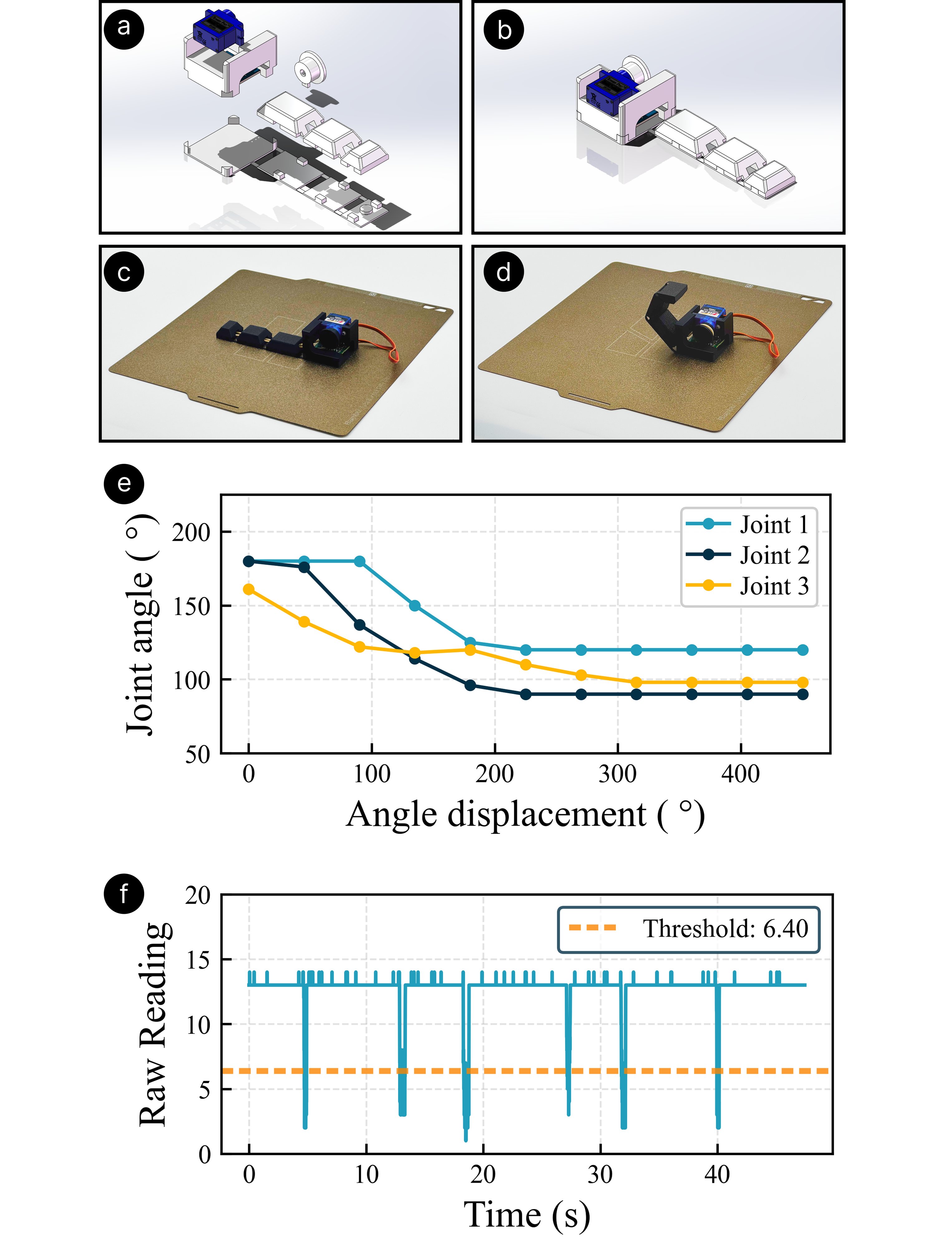} 
  \caption{Finger module demo with elastic tendon actuation. (a–b) CAD exploded and assembled views of the finger design, showing key components and assembly steps. (c–d) Printed flat-to-folded finger in extended and flexed poses. (e) Characterization of sequential joint angles during actuation compared with designed stops. (f) Capacitive touch sensor readings showing event detection at a threshold of 6.4 used to control the finger.}
  \label{fig: FingerDemoData}
\end{figure}

Capacitive touch sensing, one element of the platform I/O palette, was also implemented in the finger. Since the phalanges are printed in conductive PLA, they act as electrodes without added hardware. In our prototype, a touch event caused the servo to reverse its direction of rotation (Fig.~\ref{fig: FingerDemoData}f). In a 40 s test with six touches, all were detected and each reversed the servo, demonstrating that the same printed-electrode interface characterized in the cube can be reused in the finger module for interactive control.

\subsection{Deployable Flat Gripper}
This example demonstrates a deployable three-panel gripper built with the same method, emphasizing coordinated closing by one motor and reuse of printed electrodes for touch-based control.

\subsubsection{Self-folding}
The deployable gripper (Fig.~\ref{fig: Gripper}a to d) folds from a single flat print into a three-dimensional structure for coordinated grasping. It is fabricated as a tetrahedral net with a central triangular base and three outer rectangular panels with curved tips, joined by thin 3D-printed flexure hinges. A capstan hub at the center of the base carries three band horns spaced 120$^\circ$ apart. Each horn connects to an elastic band that anchors at a fixed hook on the corresponding panel. The three bands act in parallel: rotating the hub shortens them simultaneously and generates symmetric inward forces, producing a synchronized closing motion of all panels.

\subsubsection{Flat assembly}
\begin{figure}[ht]
  \centering
  \includegraphics[width=\columnwidth]{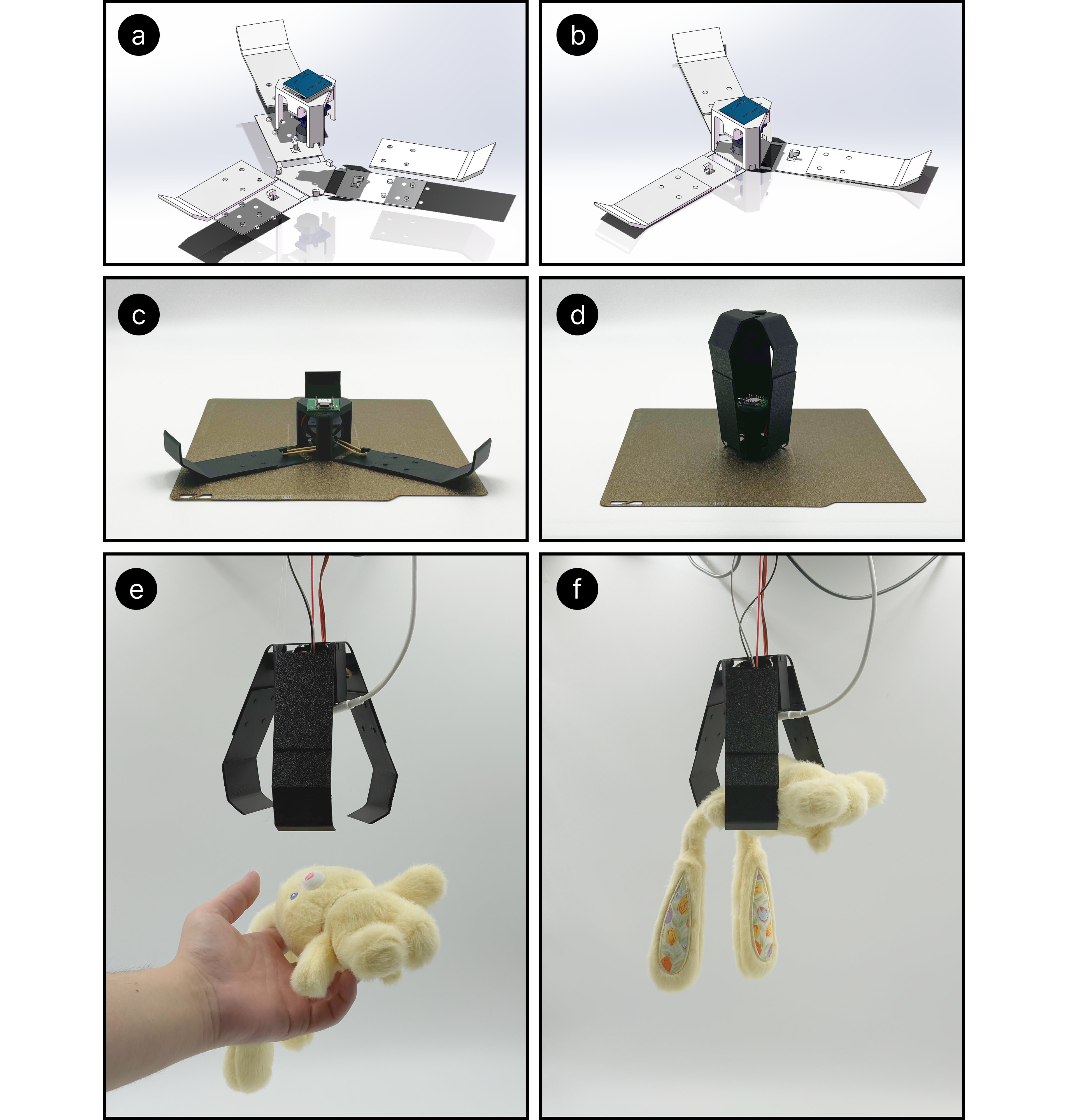} 
  \caption{Deployable flat-printed gripper demonstration. (a–b) CAD exploded and assembled views of the gripper design, showing key components and assembly steps. (c–d) Printed gripper in flat and closed states on the substrate. (e–f) Suspended operation showing relaxed and closed states, successfully grasping and lifting a 160 g toy rabbit.}
  \label{fig: Gripper}
\end{figure}

A micro servo mounted at the hub provides the single rotary input. As the servo turns the capstan, the horns retract the bands, increase their tension, and overcome hinge stiffness to close the gripper. In this way, one motor input coordinates the folding of all panels. All components, including the servo and control PCB, are installed while the device is flat to simplify assembly and ensure alignment after folding. To validate functionality, the gripper grasped and lifted a 160 g toy rabbit (Fig.~\ref{fig: Gripper}e and f). This confirms that the elastic-band-driven folding mechanism provides sufficient closing force and repeatable final geometry.

\subsubsection{Sensing via structure}
The panels are printed in conductive PLA, so the same structural faces also act as capacitive touch electrodes. The principle is the same as in the finger demo: a touch event on a panel caused the servo to reverse its direction of rotation. The sensing behavior reused the same threshold and scheme characterized in the cube and finger, enabling consistent touch responsiveness across modules.

\section{Discussion}

This work demonstrates that elastic-driven folding, combined with assembly-while-flat, provides a path to build robots that integrate structure, actuation, and sensing in a single workflow. By treating the planar sheet as the base of the foldable robot, we reduced manual effort, avoided placing parts inside enclosed cavities, and ensured consistent alignment. The approach relies only on 3D printing materials and elastic bands, making it low cost and accessible for rapid prototyping.

The hinge–band model and design map provide a way to link geometry and material parameters to the final fold angle, making it possible to choose target shapes without relying on trial-and-error. At the same time, the use of conductive PLA allows the printed sheet to serve both as structure and as sensing material, so electrodes for capacitive touch can be built into the robot during fabrication. In addition, we characterized a shared set of I/O components, including ERM motors for locomotion, magnets with Hall sensors for docking, and printed electrodes for touch. Once established at the platform level, these mappings and thresholds can be reused across modules.

The three applications illustrate how this method extends across different functions. The cube demonstrates locomotion, docking, and touch-based interaction in a form suitable for modular and swarm systems. The finger shows how an elastic tendon and printed stops enable predictable joint sequencing with a single actuator, while also supporting touch-based control through its conductive surfaces. The gripper demonstrates coordinated folding and functional grasping with one servo input. Together these applications highlight that the same fabrication and integration process can support swarm-capable robots, articulated fingers, and deployable grippers.

Several limitations remain. The hinge model assumes ideal elasticity, while real hinges suffer from fatigue, band creep, and friction. ERM locomotion is strongly surface dependent, and conductive PLA electrodes are limited by resistivity and noise. Future work should evaluate long-term reliability, expand the library of parametric seats for drop-in components, and study durability under repeated folding cycles. Multi-material printing and automated folding may further improve repeatability, robustness, and integration of multiple sensing and actuation functions.

Overall, this paper contributes a scalable and accessible route to foldable robots. By combining assembly-while-flat with elastic-driven folding and reusing a small set of platform I/O elements, we provide a generalizable workflow that enables robots with integrated actuation and sensing to be built quickly, at low cost, and across a range of applications.

\section*{Acknowledgment}
This work was supported by the Royalty Research Fund (RRF) at the University of Washington (UW).

\bibliographystyle{IEEEtran}
\bibliography{reference}

\end{document}